# Union of Experts: Adapting Hierarchical Routing to Equivalently Decomposed Transformer

Yujiao Yang, Jing Lian, *Member, IEEE*, Linhui Li, *Member, IEEE*

***Abstract*—Mixture-of-Experts (MoE) enhances model performance while maintaining computational efficiency, making it well-suited for large-scale applications. Conventional mixture-of-experts (MoE) architectures suffer from suboptimal coordination dynamics, where isolated expert operations expose the model to overfitting risks. Moreover, they have not been effectively extended to attention blocks, which limits further efficiency improvements. To tackle these issues, we propose Union-of-Experts (UoE), which decomposes the transformer model into an equivalent group of experts and applies a hierarchical routing mechanism to allocate input subspaces to specialized experts. Our approach advances MoE design with four key innovations: (1) Constructing expert groups by partitioning non-MoE models into functionally equivalent specialists (2) Developing a hierarchical routing paradigm that integrates patch-wise data selection and expert selection strategies. (3) Extending the MoE design to attention blocks. (4) Proposing a hardware-optimized parallelization scheme that exploits batched matrix multiplications for efficient expert computation. The experiments demonstrate that our UoE model surpasses Full Attention, state-of-the-art MoEs and efficient transformers in several tasks across image and natural language domains. In language modeling tasks, UoE achieves an average reduction of 2.38 in perplexity compared to the best-performing MoE method with only 76% of its FLOPs. In the Long Range Arena benchmark, it demonstrates an average score at least 0.68% higher than all comparison models, with only 50% of the FLOPs of the best MoE method. In image classification, it yields an average accuracy improvement of 1.75% over the best model while maintaining comparable FLOPs.**

## I. INTRODUCTION

Mixture of Experts (MoE) [1] is an advanced deep learning framework that effectively enhances model efficiency and strengthens its predictive capabilities. This architecture features a sophisticated gating mechanism that orchestrates a constellation of finely tuned expert networks, each excelling with distinct data subsets. With the dynamic allocation mechanism, it can enhance the model's efficiency while maintaining the optimal performance.

This work was supported by the National Natural Science Foundation of China (Nos. 52172382, 52472426). (Corresponding author: Linhui Li.)

Yujiao Yang, Jing Lian and Linhui Li are with the School of Mechanical Engineering, Dalian University of Technology, Dalian 116024, China. (e-mail: yjyang@mail.dlut.edu.cn; lianjing@dlut.edu.cn; lilinhui@dlut.edu.cn).

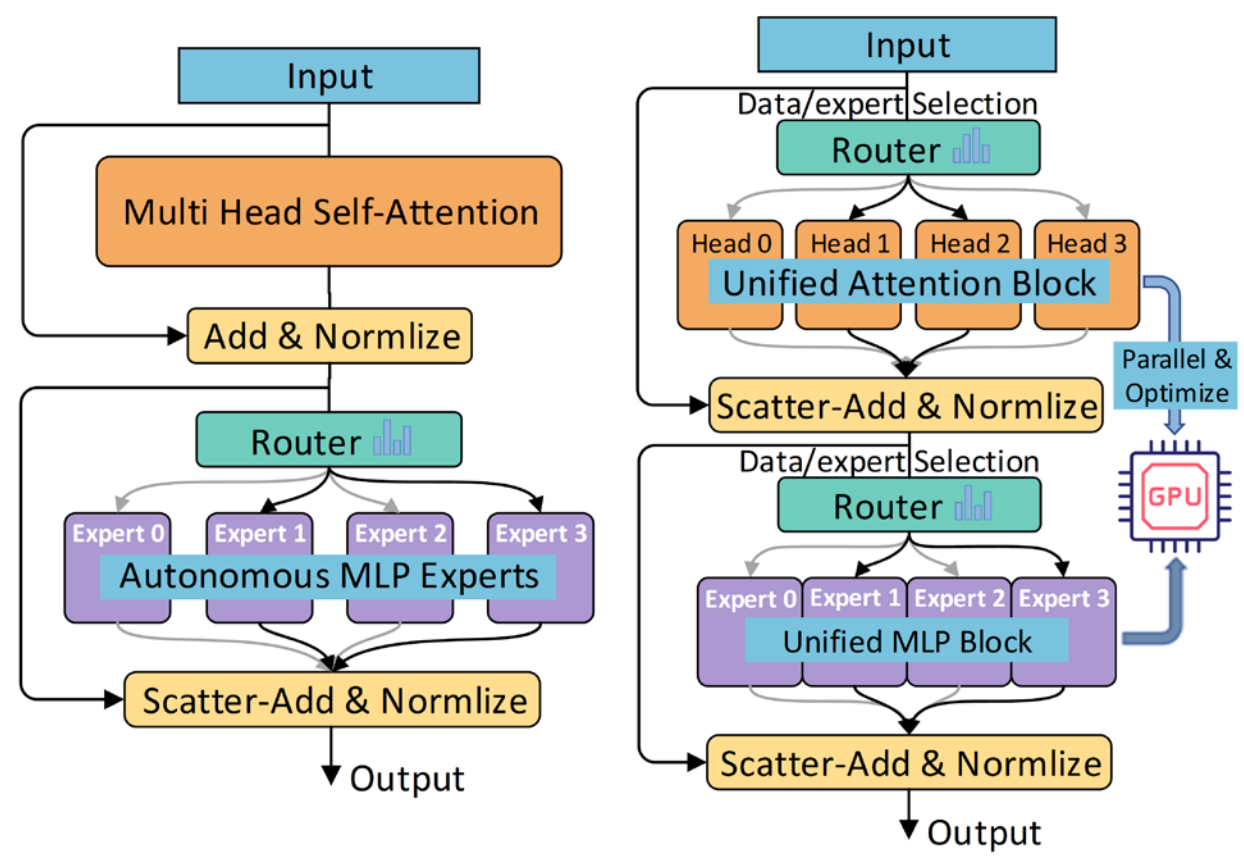

(a) Mixture-of-Experts (b) Union-of-Experts

**Fig. 1.** The architectural evolution from (a) conventional Mixture-of-Experts (MoE) framework to (b) our Union-of-Experts (UoE) framework. UoE introduce an equivalent expert partition strategy, where each expert is integrated as a specialized component of the unified model rather than an isolated sub-model. Furthermore, a hierarchical routing mechanism is applied in parallel to both attention and MLP layers, which optimizes task-specific computation paths and significantly improves computational efficiency.

Existing MoE methods are confronted with certain challenges. In terms of performance, MoE models are grounded in the concept of ensemble learning [2]. Each expert contains an independent subspace, allowing the model to learn multiple different representations in parallel across multiple subspaces [3]. However, the independent nature of experts and weighted aggregation mechanisms inherently weakens inter-expert collaborative capacity. This structural limitation introduces latent overfitting risks, which hinders effective utilization of scaling benefits in large-scale MoE architectures. In terms of efficiency, existing MoE methods are primarily applied to MLP blocks. They have not been effectively extended to attention blocks, which constrains further efficiency improvements. Moreover, they seldom succeed in achieving parallel computing, nor in optimizing efficiency based on the CUDA computing process.

To address the aforementioned issues, we propose Union-of-Experts (UoE). Fig. 1 illustrates the architectural evolution from conventional MoE to our UoE framework. The core innovation of UoE resides in its expert formation mechanism. We apply the equivalent decomposition mechanism in intra-layer model parallelism [4] to MoE architectures, constructing expert groups through holistic partitioning rather than individual aggregation. In this paradigm, each expert evolves to become one part of a whole model. The enhanced collaboration promotes collective intelligence and knowledge sharing, which in turn enhances model performance.

We also develop a group of routing mechanisms, which contain selection strategies for both the input data and experts. In the data selection strategy, we split each input sample into $m$ patches, and select no more than $c$ patches for each expert as input. In this configuration, the expert only receives one part of the sample as an input tensor, thereby effectively exploiting the locality of information while maintaining a fine-grained data routing strategy. In the expert selection strategy, we route each input sample to the top $k$ of $n$ experts. This strategy is conducive to enhancing model stability, as activated experts can observe the input sequence in a comprehensive view. Combining the two strategies allows for dynamically and precisely removing the parts of samples and experts that are not beneficial to the final results, which helps to optimize the computational efficiency while enhancing the model performance.

We applied the proposed routing mechanism to equivalent decomposed attention experts and MLP experts to build the fundamental architecture of UoE. UoE's architecture consists of Selective Multi-Head Attention (SMHA) and Union-of-MLP-Experts (UoME). SMHA maintains the multi-head design of the Multi-Head Attention (MHA) mechanism, which achieves a selective routing mechanism while enabling the model to learn diverse representations across multiple subspaces. UoME incorporates the selective routing mechanism into a decomposed MLP model, integrating the activated experts into a union similar to a larger-scale dense model. It's worth noting that the experts in both components are formulated by decomposing the MHA and MLP of the transformer model in accordance with the equivalent decomposition principle. Therefore, they are unified within the UoE architecture.

To further improve our model's efficiency, we implement parallel multi-expert computing at the algorithmic level. This approach completely resolves the issues in the parallelization of classical MoE methods. We also conduct an in-depth analysis to identify the inefficiency factors in computation. Specifically, we visualize the time cost of each operation in the complete training phase and excavate several potential inefficiency steps. By adopting the above methods, our model reduces Floating Point Operations (FLOPs) by at least 30% in comparison with the original version and achieves an average $2.26\times$ speedup and $2.68\times$ memory efficiency over the state-of-the-art MoE method. It can be anticipated that applying UoE method to larger-scale language models with a more sparser activation setting will lead to a more than 10x increase in efficiency.

We evaluate UoE on three typical tasks: language modeling, long-range sequence modeling and image classification. Experimental results demonstrate that our model consistently outperforms several methods, including the standard transformer [5], state-of-the-art MoE (including the latest DeepSeek-V3 work), and transformer variants. With a significantly reduced computational overhead, it achieves a 2.87 optimization in perplexity compared to the state-of-the-art MoE models on the WikiText-103 benchmark, and a 1.89 decrease on the One-Billion-Word language modeling benchmark. On the well-established Long Range Arena (LRA) benchmark, the average precision outperforms the state-of-the-art MoE model by 0.68%, as well as the advanced efficient transformer model by 1.9%. For the ViT-based image classification task, the average precision shows an improvement of 1.75% over the state-of-the-art MoE model. These results strongly emphasize our model's robust performance and adaptability in various domains.

## II. Related Work

### A. Variants of Transformer

Transformer [5] architecture represents a significant advancement in artificial intelligence, bringing forth unparalleled capabilities alongside the challenge of resource-intensive training and serving processes. Significant enthusiasm has been ignited for actively enhancing the efficiency of transformer architectures. A preponderance of the works focuses on reducing memory usage and improving computational efficiency. Sparse Transformer [6] introduces sparse attention mechanisms to selectively attend to relevant tokens within a sequence. Linformer [7] and Lightning-Attention [8, 9] leverage linear-complexity self-attention mechanisms, catering to large-scale data and lengthy sequences. Reformer [10] mitigates memory constraints by employing reversible layers and locality-sensitive hashing techniques. Flash Attention [11] reduces memory access costs through tiling, while its subsequent version [12] further enhances performance by optimizing memory access and computation fusion. DeepSeek-V2 [13] introduces the Multi-Head Latent Attention (MLA) mechanism, which employs low-rank joint compression to enhance training efficiency and reduce the KV cache size during inference. Tensor Product Attention (TPA) [14] dynamically constructs QKV as context-dependent decomposed tensors, enabling adaptive adjustments and facilitating seamless integration with effective Rotary Position Embedding.

The other works aim to enhance the modeling capabilities of long sequences. Transformer-XL [15] introduces segment-level recurrence to enhance sequence modeling beyond the scope of the original Transformer. Sinkhorn Transformer [16] fuses the Sinkhorn algorithm with self-attention mechanisms to improve sequence modeling accuracy. Long-Short-Term Memory Transformer [17] combines the Transformer with Long Short-Term Memory (LSTM) mechanisms and thus enhances the modeling capability for long sequences. SeerAttention [18] integrates a learnable gating mechanism into the standard attention mechanism, enabling adaptive selection of salient blocks within the attention map. Although these innovations have notably improved the performance of the transformer frameworks in their respective domains, they are typically model-specific and may not be universally applicable. By comparison, our work provides a model-free

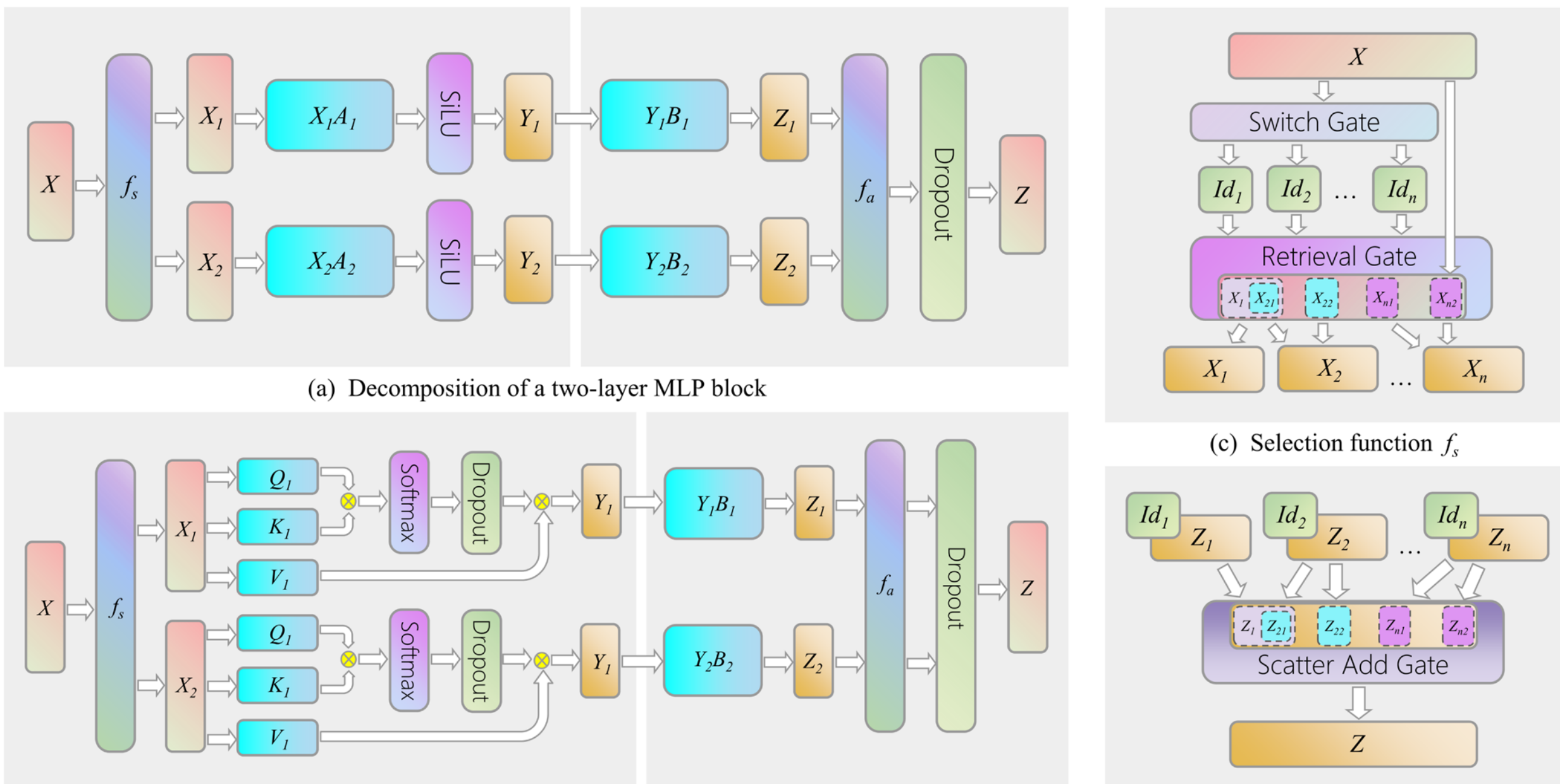

(a) Decomposition of a two-layer MLP block

(b) Decomposition of a Multi-Head Attention block

(c) Selection function $f_s$

(d) Aggregation function $f_a$

**Fig. 2.** The concrete architecture of UoE. We equivalently decompose the MLP block and the attention block of the transformer into $n$ independent computation branches (with $n = 2$ in the figure), and regard each branch as an expert, as shown in panels (a) and (b). We selectively route the input batch to each expert for independent calculation and aggregate the results of experts based on the routing indices. Panels (c) and (d) illustrate the mechanisms of the selection function and the aggregation function, where $X_{ij}$ can be either a patch or a sample, corresponding to data selection and expert selection respectively.

method. By simply adding a selection mechanism to existing model parallel transformers, we can train a model more efficiently while maintaining or even improving its performance.

*B. Mix of Experts*

Mixture-of-Experts (MoE) enhances model capacity while maintaining the computation overhead, thereby attaining superior performance relative to dense models across various domains. In classical MoE layers, a group of collaborative experts works in unison to address complex tasks, while each can be a simple Feed-Forward Network or a fully independent submodel [21]. To effectively manage this ensemble of experts, MoE introduces routing layers that decide which experts to activate based on the input, and aggregation layers that combine the outputs into a unified response.

Extensive works have emerged in the development of MoE architecture. The first successful deep learning-based MoE [1] inserted a routing layer between two LSTM layers to select a sparse combination of activated experts, reaching state-of-the-art performance in machine translation. Despite this success, however, follow-on research was relatively dormant with greater emphasis on directly studying the Transformer [5].

This changed with the release of GShard [22] and Switch Transformers [23], both of which replaced the feed-forward layers in transformer architecture with expert layers. While the experts-as-a-layer approach has become the dominant paradigm, more recent works revisit the concept of experts as fully independent models [24, 25], which confers a benefit of modularity and composability. Skywork-MoE [26] introduces innovative gated logit normalization and adaptive auxiliary loss coefficients to enhance expert diversity and training efficiency. DeepSeek-MoE [27] employs fine-grained common experts and shared experts, which enhance expert specialization across different tasks.

Although the existing works have made significant strides, they are constrained by the organizational form of experts. In MoE frameworks, each expert works as an independent individual. The effectiveness of this pattern has been empirically validated, showcasing its reliability in practical scenarios. However, such a design undermines the collaboration ability of experts, which could potentially impair the model's ability to capture complex patterns. Moreover, it may introduce additional gather layers, leading to increased computational demands. In contrast, our approach sets one part of a whole model as an expert, enabling a highly collaborative group of experts from a global perspective, without adding any fusion structure. It is worth noting that our approach is not aimed at replacing all groups of experts with union models, but rather at introducing a selection mechanism without damaging the model's original organizational form, no matter if it was originally a dense model or a mixture of sub-models or heads (such as a Multi-Head Attention layer).

*C. Model Parallelism*

Model parallelism is a computational strategy designed to distribute the workload of large neural network models across multiple processing units [28]. There are generally two types of parallelism: tensor parallelism and pipeline parallelism. Pipeline parallelism offers an alternative by dividing the model into distinct stages, each handled by a different device. This approach was notably advanced by the GPipe framework [29], which implemented synchronous mini-batch pipelining to train models with billions of parameters. On the other hand, tensor parallelism partitions a tensor operation (such as matrix-matrix multiplication) across multiple devices to accelerate computation or increase model size. In the field, Megatron-LM [4] demonstrated the potential of tensor parallelism in scaling transformer models. By splitting the weight matrices and activations across GPUs, it achieved substantial improvements in training large language models. We build our approach upon the work of Megatron-LM. Specifically, we apply its lossless weight splitting algorithm to equivalently decompose a whole model into a group of UoE experts. Compared to Megatron-LM, our approach only needs to activate a subset of weights, thus can dynamically eliminate unnecessary computation overhead.

## III. IMPLEMENTING UNION-OF-EXPERTS

We construct the Union-of-Experts (UoE) architecture based on the standard transformer. The architecture of the UoE model is depicted in Fig. 2. We structurally decompose the model into a parameterized set of experts while preserving functional equivalence, and integrate a hierarchical routing strategy that dynamically routes input subspaces to activated experts via sparse gating operations. We will introduce the overall model design in this section. The details of the decomposition method can be found in Appendix A.

### *A. Selective Multi-Head Attention*

In the UoE architecture, the attention block is developed based on the standard Multi-Head-Attention (MHA) block. Let $n_a$ denote the number of attention heads, which is equal to the number of experts, $l$ denote the sequence length, $d$ denote the embedding dimension, $d_h$ denote the dimension per head or expert, and $h \in \mathbb{R}^{l \times d}$ denote the input sample at a UoE attention block. We apply a data selection function on the input $h$ (see Appendix B for details), which can be formulated as follows:

$$g = \text{softmax}(\text{FFN}(h)) \tag{1}$$

$$id'_j = \text{Topk}(\{g_{i,j} \,|1 \le i \le n_a\}, k) \tag{2}$$

$$l_a = \max_{i=1}^{n}(\sum_{j=1}^{m} I(i \in id'_j)) \tag{3}$$

$$id_i = \text{Topk}(\{g_{i,j} \,|1 \le j \le l\}, l_a) \tag{4}$$

$$h'_i = f_S(h, id_i) \tag{5}$$

Where $\text{FFN}(\cdot)$ is the feed-forward network, $\text{Topk}(\cdot, k)$ denotes the set of the $k$ highest affinity scores, $I$ denotes the indicator function, $f_S(\cdot)$ refers to an index selection function available in the pytorch API. $g \in \mathbb{R}^{n \times m}$ denotes the gate value, $l_a$ denote the maximum length of processed inputs $h'$, which is equal to the maximum patch counts multiplied by the number of patches in one head, $id \in \mathbb{R}^{n_a \times l_a}$ denotes the routing indices, $h' \in \mathbb{R}^{n_a \times l_a \times d}$. We apply three column parallel linear functions to $h'$, to calculate the query $q$, key $k$, and value $v$:

$$\begin{aligned}[q_1, q_2, ..., q_n] = q = h' \odot W^q \\ = [h'_1 W_1^q, h'_2 W_2^q, ... h'_n W_n^q]\end{aligned} \tag{6}$$

$$[k_1, k_2, ..., k_n] = k = h' \odot W^k \tag{7}$$

$$[v_1, v_2, ..., v_n] = v = h' \odot W^v \tag{8}$$

Where $W^q, W^k, W^v \in \mathbb{R}^{n_a \times d \times d_h}$ denotes the projection matrices, $q, k, v \in \mathbb{R}^{n_a \times l_a \times d_h}$. To improve the model's extrapolation ability, we integrated Rotary Position Embedding (RoPE) into the query and key:

$$q_i = [q_i^c, q_i^r], q_i{}' = [q_i^c, \text{RoPE}(q_i^r)] \tag{9}$$

$$k_i = [k_i^c, k_i^r], k_i{}' = [k_i^c, \text{RoPE}(k_i^r)] \tag{10}$$

Where $q^c \in \mathbb{R}^{n_a \times l_a \times d_{qc}}$, $k^c \in \mathbb{R}^{n_a \times l_a \times d_{kc}}$ denotes the constant part in query and key, $q^r \in \mathbb{R}^{n_a \times l_a \times d_{qr}}$, $k^r \in \mathbb{R}^{n_a \times l_a \times d_{kr}}$ denotes the part where RoPE will be applied, $d_{qc} + d_{qr} = d_{kc} + d_{kr} = d_h$, $1 \le i \le n_a$. We combine the attention queries $q'$, keys $k'$, and values $v$ to calculate the attention output:

$$s_i = \text{Softmax}(\frac{q_i{}'^T k_i{}'}{\sqrt{d_h}}) \tag{11}$$

$$o_i = s_i v_i \tag{12}$$

$$u' = o \odot W^o = [o_1 W_1^o, o_2 W_2^o, ..., o_n W_n^o] \tag{13}$$

Where $W^o \in \mathbb{R}^{n_a \times d_h \times d}$ denotes the output projection matrix. Ultimately, we apply the index add function on $u'$ to yield the final output $u$:

$$u_j = h_j + \sum_{i=1}^{n} \sum_{k=0}^{l_a} u'_{i,t} \cdot I(id_{i,t} = j) \tag{14}$$

### *B. Union-of-MLP-Experts*

In the design of the UoE style multilayer perceptron block, we incorporated lossless model decomposition and expert routing methods into a two-layer MLP. Let $n_m$ denote the number of MLP experts, $l_m$ denote the maximum length of processed inputs $u'$, $d_e$ denote the project dimension of experts. Given a sample $u \in \mathbb{R}^{l_{inp} \times d}$ in an input batch $H$, the data processing procedure can be formulated by a selection function $f$ defined in Equations 1 to 5:

$$u', id^u = f(u) \tag{15}$$

While $id \in \mathbb{R}^{n_m \times l_m}$ denotes the routing indices, $u' \in \mathbb{R}^{n_m \times l_m \times d}$. We implement column partition on the first MLP layer, and row partition on the second one to preserve the equivalence before and after partitioning, see Appendix A for more details. We can outline the expert processing procedure in the following manner:

$$\begin{aligned}[v_1, v_2, ... v_n] = v = \varphi(u' \odot A^I) \\ = [\varphi(u'_1 A_1^I), \varphi(u'_2 A_2^I), ... \varphi(u'_n A_n^I)]\end{aligned} \tag{16}$$

$$[h'_1, h'_2, ... h'_n] = h' = v \odot A^O \tag{17}$$

Where $A^I \in \mathbb{R}^{n_m \times d \times d_e}$ denotes the up-projection matrix, and $A^O \in \mathbb{R}^{n_m \times d_e \times d}$ denotes the down-projection matrix. Finally, we use the index add function to add each element in sequence $h'$ to the specified position in the input $u$ to obtain the output $h$:

$$h_j = \varphi\left(u_j + \sum_{i=1}^{n}\sum_{t=0}^{l_m} h'_{i,t} \cdot I(id_{i,t} = j)\right) \tag{18}$$

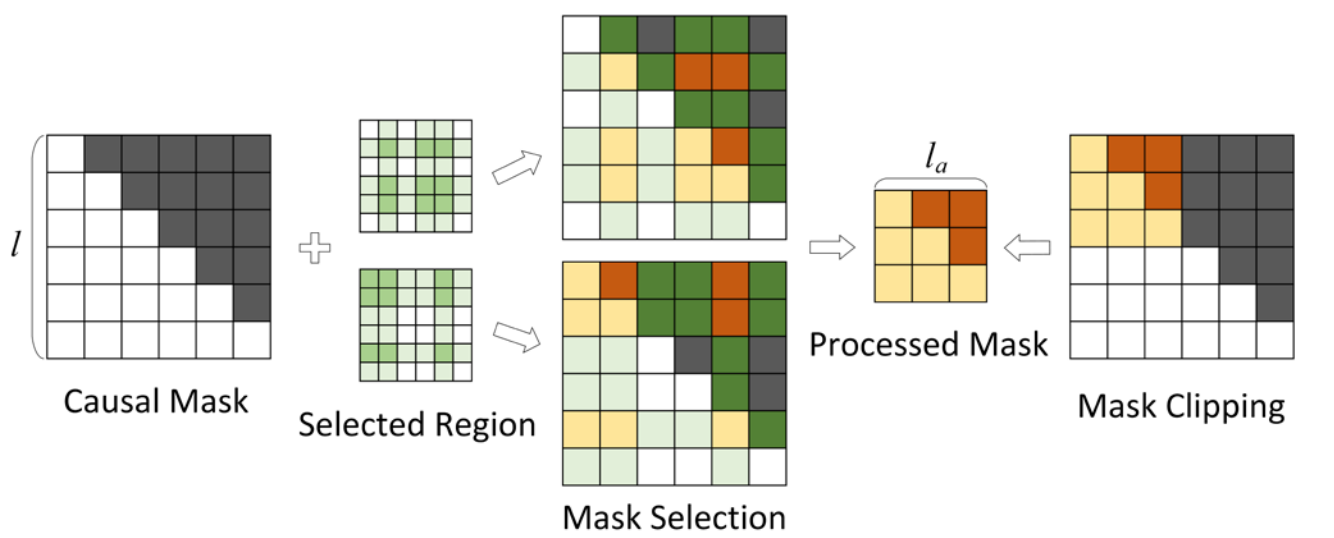

**Fig. 3.** The processing of casual mask in data selection paradigm. Each expert requires the computation of a casual mask based on its corresponding routing indices. In self-attention mechanism, we apply the same indices on both queries and keys. As a result, we obtain an easily producible $l_a \times l_a$ lower triangle matrix independent of the indices, which is applicable to all experts.

Where $\varphi(\cdot)$ denotes the SiLU activation function. It's worth noting that in our method, the nonlinear activation function is positioned after the addition operation. This arrangement allows for the synthesis of activated experts into an integrated large - scale MLP computational graph.

*C. Attention Mask*

When integrating data selection into the mask attention mechanism, it is necessary to select elements from the original attention mask to form a new group of masks that match with the group of selected data. For the padding mask $m^p \in \mathbb{R}^l$, we use the indices to perform indexing along the sequence dimension:

$$m_i^{p'} = f_S(m^p, id_i) \tag{19}$$

Where $f_S(\cdot)$ refers to the index selection function, $m_i^{p'} \in \mathbb{R}^{l_a}$ denotes the padding mask corresponding to the i-th expert's input. With regard to the causal mask $m^c \in \mathbb{R}^{l \times l}$, it is required to perform indexing operations along two dimensions for $n_a$ times. Considering that $m^c$ and the input sample $h \in \mathbb{R}^{l \times d}$ are comparable in size, this process will lead to a non-negligible computational overhead. Here we introduce an efficient and equivalent solution. As illustrated in Fig. 3, because of the symmetry between the two indexing operations, the generated mask is essentially independent of its position. Thus, we take the $l_a$-th order principal matrix of the original mask as the shared sequence mask $m^{s'} \in \mathbb{R}^{l_a \times l_a}$ for all experts' inputs, enabling the model to inherit the autoregressive capability of the transformer.

*D. Load Balancing Loss*

Prior research indicates that the imbalanced expert load in MoE models may result in routing collapse and computational inefficiency in expert parallelism scenarios [30]. To mitigate excessive imbalance within the given sequence, we employ an sequence-wise load balancing loss similar to Dai et al. [27]:

$$\mathcal{L}_{\text{Bal}} = \alpha \sum_{i=1}^{n} f_i P_i \tag{20}$$

$$f_i = \frac{n}{kl}\sum_{t=1}^{l} I\left(g_{i,t} \in \text{Topk}\left(\left\{g_{j,t} \middle| 1 \le j \le n\right\}, k\right)\right) \tag{21}$$

$$P_i = \frac{1}{l}\sum_{t=1}^{l} g_{i,t} \tag{22}$$

Where $\alpha$ denotes a hyper-parameter named balance factor, $n$ denotes the number of experts in attention or MLP block, $k$ denotes the number of activated experts, and $g$ denotes the gating value. The auxiliary load balancing loss promotes expert load distribution for each sequence to be balanced.

*E. Efficiency Optimization*

To improve UoE's efficiency, we perform an analytical review of potential inefficiencies and implement apposite improvements. In the preprocessing phase, we calculate the gating value $g$ to dynamically route data subsets into several experts. In the postprocessing phase, we use $g$ to route expert outputs to their correct location. Existing methods ordinarily employ the routing procedure serially through an n-step for loop. With additional memory copy operations such as Memcpy DtoD, it exacerbates the time complexity. To tackle this issue, we uniformly encode all data indexes, and retrieve all experts' inputs within one index selection operation. Similarly, we apply only one scatter add operation to achieve the inverse process of the retrieve operation. In this way, only one memory copy operation is needed in each phase, leading to a significant reduction in time costs.

In the propagation phase, matrix operation accounts for the majority of the computational overhead. Existing MoE methods have difficulties in parallel computing, leading to a significant reduction in computational efficiency. For instance, we observed that the non-parallelized implementation of MoEs sometimes even exhibited a higher time cost than a dense model with the same configuration in a scenario where only 50% of the parameters of MoEs are activated. From another perspective, matrix operation processes involve three procedures: operand transmission, computation, and result writing back. Performance profiles of various MoE implementations indicate that each of these factors has the potential to become a performance bottleneck.

In light of the aforementioned considerations, we optimized the matrix operation process. We transformed inputs and weights into tensors with matching dimensions, and then performed batch matrix operations. We conducted this procedure using the baddbmm function, as it effectively reduces computational overhead through three factors: 1. The function exhibits reduced time costs when transferring learnable parameters to CUDA cores. 2. The function implements an operation that integrates matrix multiplication and bias addition, with execution times comparable to single multiplication. 3. The function performs an in-place operation that directly updates the bias terms with the computed results, thereby eliminating the computational overhead associated with a secondary write operation. Overall, by introducing the above optimizations, our model achieves a speedup of over 30% compared to the original version.

## IV. Experiments

In this section, we assess the performance of our proposed UoE method through a series of experiments conducted in both language and vision domains. In detail, we designed three experiments: language modeling, long-range sequence modeling, and image classification, each focusing on a specific challenge.

### *A. Natural Language Modeling*

**Benchmark Description.** In the natural language modeling experiments, we train and evaluate our model on two datasets: Wikitext-103 and One Billion Word.

WikiText-103 is a large-scale language modeling dataset derived from Wikipedia, consisting of 103M training tokens from 28K articles, with an average length of 3.6K tokens per article, which allows testing the ability of models to capture long-term dependencies and contextual relationships within extended text sequences.

One Billion Word, on the other hand, is a corpus containing approximately 1 billion words of diverse text sourced from news articles, serving as a robust benchmark for large-scale language modeling with a focus on handling diverse and noisy real-world data.

**Comparison Baseline.** We compare our method with state-of-the-art MoE methods, including DeepSeek-MoE [27], DeepSeek-V3 [31], Skywork-MoE [26] and XMoE [32]. DeepSeek-MoE employs fine-grained expert specialization for task allocation within specialized domains and utilizes shared experts to enhance knowledge transfer across different tasks. Its successor version, DeepSeek-V2 [13] and DeepSeek-V3 introduce a new Multi-Head Latent Attention (MLA) mechanism, which compresses key-value pairs into latent representations for efficiency optimization. Given that DeepSeek-V2 and V3 only exhibit subtle differences in routing method (i.e., the V3 replaces the softmax activation with the tanh activation), we select the latest V3 version as a representative comparison model. Skywork-MoE refines expert activation selection through gating logit normalization and employs adaptive auxiliary loss coefficients to dynamically regulate regularization in response to expert load balance. XMoE employs fine-grained experts and a threshold-based router to enhance model adaptability and promote sparsity in large-scale models. We also incorporate Transformer [5, 33] as a fully activated baseline model to facilitate a more thorough evaluation of efficiency.

**Evaluation Metrics.** To evaluate the performance of our model on the language modeling task, we adopt Perplexity (PPL) and Floating Point Operations (FLOPs) as the primary evaluation metrics. Perplexity is a metric that measures how well a language model predicts a sequence of words:

$$PPL = \exp\left(-\frac{1}{N}\sum_{i=1}^{N}\log P(w_i \,|\, w_{1:i-1})\right) \tag{23}$$

Where $N$ is the total number of words in the target dataset, $P(w_i \,|\, w_{1:i-1})$ is the predicted probability of the *i*-th word given the previous context $w_i \,|\, w_{1:i-1}$.

The FLOPs provide a hardware-agnostic estimate of computational complexity, which is particularly relevant for comparing models in terms of efficiency. As the FLOPs calculation methods differ across algorithms, we utilized calflops [34] to calculate FLOPs automatically.

**Implementation Details.** We developed an experimental framework with reference to the code implementation of Dai et al. [15]. The parameters of each model were configured based on the principle of approximate model size and floating-point operations. Specifically, we adopted the transformer configuration from [15] as the base configuration, while the remaining baseline models shared the same common parameters (e.g., Number of Layers, Attention Heads, and dimensions). For the parameters unique to the baseline model, we referred to the original code implementation for configuration. With regard to each MoE model, we activated half of the MLP experts to process each input. This configuration ensures sufficient expert participation while preventing overfitting, which helps to balance loads and enhance efficiency. Similarly, for DeepSeek-V2, which uses a Multi-Head Latent Attention mechanism, we set its lora rank equal to half of the embedding dimension. All of the experiments employed the same training settings, such as batch size, sequence length, and training steps.

In terms of UoE's configuration, we applied the data selection paradigm to attention blocks and the expert selection paradigm to MLP blocks. We set the number of MLP experts equal to that in the baselines and activated a half for each input sample. In the Multi-Head Attention block, we set the number of experts equal to the number of heads and assigned each expert 50% of the input sequence. In this configuration, the theoretical FLOPs required by our model is roughly 50% of the count associated with the transformer baseline. Experimental results are depicted in Table 1.

**Results.** Table 1 illustrates the comparative results. The performance metrics indicate that our model noticeably outperforms all state-of-the-art baselines, which highlights the advantages of overall expert collaboration and fine-grained collaborative routing. We achieved a test perplexity of 24.09/ 24.52 in the wikitext103/One Billion Word datasets, which is superior to that of the Transformer (24.23/24.70) and all competing MoE-based works (best: 26.96/26.41 from DeepSeek-V3).

Prevalent MoE methods are derivative models that replace the MLP block in the transformer with a Mixture-of-Experts architecture. Among them, some works focus on introducing fine-grained routing (i.e., DeepSeek-MoE and XMoE). Other research efforts emphasize enhancing routing precision and balance (i.e., Skywork-MoE). By comparison, our approach not only possesses these characteristics but also enhances expert mixing methods to promote in-depth expert collaboration, leading to significantly improved performance compared to the above methods. Furthermore, there is a class of MoE methods that incorporates an optimization scheme particularly for the attention block, e.g., the Multi-Head Latent Attention mechanism introduced in DeepSeek-V3 and our UoE attention block. Experimental results demonstrate that under similar FLOPs metrics, our approach achieves superior performance compared to DeepSeek-V3 (24.09/24.52 vs 26.96/26.41), which underscores the efficacy of our approach, especially with respect to our Union-of-Experts based attention mechanism.

TABLE I
PERFORMANCE COMPARISON BETWEEN UOE AND BASELINES ON WIKITEXT-103 AND ONE BILLION WORD BENCHMARKS.

| Model | Wikitext-103 | | One Billion Word | |
|---|---|---|---|---|
| | PPL | TFLOPs | PPL | TFLOPs |
| Transformer[5] | 24.23 | 2.67 | 24.70 | 6.27 |
| XMoE[32] | 30.60 | 2.67 | 29.01 | 4.69 |
| DeepSeek-MoE[27] | 28.30 | 2.30 | 28.83 | 5.46 |
| Skywork-MoE[26] | 29.10 | 2.42 | 28.65 | 5.54 |
| DeepSeek-V3[31] | 26.96 | 2.65 | 26.41 | 5.26 |
| Ours | **24.09** | **1.74** | **24.52** | **4.53** |

We also analyzed the FLOPs of experimental models. As shown in Table 1, the FLOPs of our model in Wikitext-103/One Billion Word (1.74/4.53) are better than all competing models, including DeepSeek-V3 (2.65/5.26), Skywork-MoE (2.42/5.54), DeepSeek-MoE (2.30/5.46), XMoE (2.67/4.69), and Transformer (2.67/6.27). The model's FLOPs are dependent on its architectural design and parameter configuration. In our experiment, MLP blocks of each MoE method activate only half of their parameters. With regard to attention blocks, unlike other works (e.g., DeepSeek-MoE) that utilize softmax attention, our UoE design has a partially activated attention mechanism, while DeepSeek-V3 employs a low-rank adaptive one; both contribute to FLOPs reduction. On the other hand, in the parallel implementation of selective activation, the imbalance in expert workloads will result in additional costs, which leads to higher actual FLOPs than the theoretical value (i.e., 50% FLOPs of the Transformer). Nonetheless, our UoE still maintains FLOPs superior to leading methods.

*B. Context Modeling on Long Range Arena Benchmark*

**Benchmark Description.** To evaluate UoE's long text modeling capability, we trained our models on the widely recognized Long Range Arena (LRA) benchmark [35]. This benchmark comprises several sub-tasks, which evaluate a model's performance from various perspectives, including compositionality, hierarchical structure, and spatial reasoning.

ListOps. The Long ListOps task tests the model's ability to handle hierarchically structured data within extended contexts, using sequences of up to 2K in length.

Text. The Byte-Level Text Classification task evaluates the model's capacity to classify documents from byte-level data, simulating real-world tasks like spam detection.

Retrieval. The Byte-level Document Retrieval task assesses a model's ability to encode and retrieve documents based on similarity scores.

Image. The Image Classification on Sequences of Pixels task involves classifying images represented as 1D sequences, requiring models to capture spatial dependencies.

Pathfinder. The Pathfinder task tests long-range spatial dependencies in visual data, requiring models to determine whether two points are connected by a path.

**Comparison Baseline.** Following the same protocol as the previous experiment, we compare our method with the Transformer and state-of-the-art MoE methods, which include DeepSeek-MoE, DeepSeek-V3, Skywork-MoE, and XMoE, following the same protocol as in the previous experimental design. Considering that our model can be regarded as a variant of the transformer architecture, we also select several efficient transformer variants, including Reformer [10], Linformer [7], Performer [36], Linear Attention [37], Nyströmformer [38] and Flash-Attention [11], as our baseline models. Reformer introduces locality-sensitive hashing to approximate attention mechanisms, which helps to reduce time complexity from quadratic to linear. Linformer employs low-rank projections to approximate the full attention matrix. Performer utilizes kernel methods to approximate softmax attention. Linear Attention replaces the softmax-based attention mechanism with a linear transformation. Nyströmformer applies Nyström approximation to the attention mechanism to reduce computation complexity. Flash Attention leverages optimized memory access, block-wise computation, and efficient parallelization, achieving faster execution and reduced memory usage.

**Evaluation Metrics.** We adopt widely used Accuracy and FLOPs as evaluation criteria for the classification task. Accuracy measures the proportion of correctly classified instances in the dataset, providing a straightforward assessment of the model's performance. The FLOPs is employed to test the computational efficiency of the model by measuring the total number of floating-point operations needed. In this experiment, we provide the average ratio of tested model FLOPs to standard transformer FLOPs in each subtask of the LRA benchmark. In general, the metrics provide a balanced evaluation of the model's performance and efficiency.

**Implementation Details.** We designed and trained our models based on the frameworks established by Tay et al. [35] For a fair comparison, we follow the experimental setting for efficient transformers as used in Xiong et al. [38] and Zhu et al. [39]. For some works not covered, we provide the best results from Dao et al. [11]. In terms of the MoE comparison model, we adopted a similar parameter configuration to that used in the previous experiment. (e.g., set the activate ratio to 0.5 and MLA's lora rank equal to half of the embedding dimension). For consistency, the same training parameters are applied in all experiments.

In terms of UoE's setup, considering that in LRA datasets, the input sequences are notably long, and the information within these long sequences is distributed sparsely, we apply the data selection paradigm on both attention blocks and MLP blocks of our model to effectively extract the valuable information embedded at various locations within sequences. Similar to the previous configuration, we set the number of MLP experts equal to that in the baselines, while in the Multi-Head Attention Block, it is equal to the attention head count. For all experts in the data selection paradigm, the activation ratio is set to 0.5. Experimental results are depicted in Table 2.

TABLE II
THE EXPERIMENTAL RESULTS ON LONG RANGE ARENA (LRA) BENCHMARK

| Task | ListOps | Text | Retrieval | Image | Pathfinder | Average | FLOPs |
|---|---|---|---|---|---|---|---|
| Efficient Transformers | | | | | | | |
| Transformer [5] | 37.13 | 65.35 | 82.30 | 42.44 | 74.16 | 60.28 | 1.00× |
| Reformer [10] | 36.44 | 64.88 | 78.64 | 43.29 | 69.36 | 58.52 | 0.19× |
| Linformer [7] | 37.38 | 56.12 | 79.37 | 38.56 | **76.34** | 57.55 | 0.23× |
| Performer [36] | 36.80 | 63.60 | 82.20 | 42.10 | 69.90 | 58.92 | 0.23× |
| Linear Attention [37] | 38.80 | 63.20 | 80.70 | 42.60 | 72.50 | 59.56 | 0.18× |
| Nyströmformer [38] | 37.34 | **65.75** | 81.29 | 41.58 | 70.94 | 59.38 | 0.32× |
| Flash Attention [11] | 37.60 | 63.90 | 81.40 | 43.50 | 72.70 | 59.82 | **0.17**× |
| Mix-of-Experts Methods | | | | | | | |
| XMoE [32] | 38.00 | 63.42 | 81.33 | 42.10 | 73.65 | 59.70 | 0.99× |
| DeepSeek-MoE [27] | 37.80 | 65.12 | 81.35 | 43.55 | 74.08 | 60.38 | 0.99× |
| SkyworkMoE [26] | 37.85 | 65.18 | 82.25 | 45.24 | 74.11 | 60.93 | 0.99× |
| DeepSeek-V3 [31] | 38.26 | 65.50 | 81.77 | 45.15 | 74.52 | 61.04 | 0.74× |
| UoE (ours) | **38.91** | 65.61 | **82.82** | **46.09** | 75.17 | **61.72** | 0.37× |

**Results.** Comparisons with models on the LRA benchmark are shown in Table 2. In the ListOps task, which models hierarchically structured data, our UoE model achieved a significant accuracy of 38.91, outperforming all competing efficient transformer models (best: 38.80 from Linear Attention), as well as all MoE methods (best: 38.26 from DeepSeek-V3), showing its capacity for understanding recursive patterns and operating over long-context sequences. In the Retrieval task, which evaluates the model's encoding and storage of compressed representations, our model achieved an accuracy of 82.82, ranking among the top contenders of transformer variants (best: 82.20 from Performer) and MoEs (best: 82.25 from Skywork-MoE). Our model also highlights its proficiency in image processing. The performance of our model in the Image task was measured at 46.09, exceeding the performance of all competing models, including the previous highest score of 45.24 reported by Skywork-MoE.

The remaining tasks further demonstrate the superiority of our model. Achieving an accuracy of 65.61 on the Text task and 75.17 on the Pathfinder task, our model significantly outperformed all MoE methods, with the highest MoE accuracy being 65.50/74.52 from DeepSeek-V3. It is also competitive with advanced efficient transformers. In the Text task, the best score is 65.75 achieved by Nyströmformer, while in the Pathfinder task, it is 76.34 attained by Linformer. Overall, our model achieves an average score of 61.72 with 37% of the FLOPs of the standard transformer model. In comparison to the latest state-of-the-art DeepSeek-V3 method, our model achieves an average score improvement of 0.68% while requiring only 50% of its computational costs. This performance is also on par with the advanced Flash Attention, which achieves an average score of 59.82 with 17% of the FLOPs. Note that as a variant of MoE, UoE's complexity is determined by the application context. When constructed with a linear complexity attention mechanism such as Flash Attention, it can achieve significantly greater efficiency improvements.

The results presented above demonstrate the superiority of our proposed method. By introducing the unique Union-of-Experts mechanism in the entire transformer, our model overcomes the inherent limitations of traditional MoE methods and the transformer architecture, surpassing existing state-of-the-art MoE methods. Moreover, even without an explicit efficient architecture design, our model's performance is competitive with that of the state-of-the-art efficient transformer model. In reality, our model exhibits a natural advantage in modeling long sequences. On the one hand, each expert in our model only receives a portion of patches or tokens from the sequence as input. The expert may be subjected to a finite set of inputs that are positioned at relatively distant intervals along the sequence, which indicates its efficiency in handling long-range dependencies. On the other hand, the expert is able to focus on a specific input region, thereby facilitating the effective processing of local information. By applying this paradigm to the attention block and the MLP block, we have implemented the extraction and specialization of useful knowledge from the sequence, significantly enhancing the model's ability to handle long-range dependencies. In addition, the FLOPs of XMoE, DeepSeek-MoE, and Skywork-MoE show no significant differences from that of the standard transformer. This can be attributed to the quadratic complexity of Multi-Head Attention. For long sequence inputs, the computational overhead of Multi-Head Attention constitutes the major portion. As these models primarily optimize the MLP, their efficiency remains largely unaffected.

TABLE III
EXPERIMENTAL RESULTS ON VIT-BASED IMAGE CLASSIFICATION TASK

| Model | Accuracy | | GFLOPs |
|---|---|---|---|
| | Cifar-10 | Cifar-100 | |
| Transformer [5] | 74.45 | 42.60 | 340.21 |
| XMoE [32] | 70.90 | 41.27 | 264.99 |
| DeepSeek-MoE [27] | 72.31 | 40.82 | 245.08 |
| SkyworkMoE [26] | 80.87 | 49.25 | 281.83 |
| DeepSeek-V3 [31] | 81.51 | 53.16 | **210.38** |
| UoE-DD (ours) | 81.81 | 55.09 | 220.99 |
| UoE (ours) | **82.70** | **55.47** | 220.34 |

### *C. Image Classification*

**Dataset Description.** We perform the image classification experiments on the CIFAR-10 and CIFAR-100 datasets.

CIFAR-10 consists of 60,000 32x32 color images categorized into 10 classes, with 6,000 images per class, split into 50,000 training images and 10,000 test images. Each image in CIFAR-10 is labeled with a single class, representing common objects such as airplanes, automobiles, etc.

CIFAR-100 is an extension of CIFAR-10, containing 60,000 32x32 color images divided into 100 fine-grained classes, which are grouped into 20 coarser super-classes. Each class in CIFAR-100 contains 600 images, with 500 used for training and 100 for testing. This dataset is more challenging due to its larger number of classes, hierarchical labeling with fine-grained and coarse labels, and smaller amount of training data.

**Comparison Baseline.** Analogous to language modeling experiments, we compare our method with the Transformer and state-of-the-art MoE methods, which include DeepSeek-MoE, DeepSeek-V2, Skywork-MoE, and XMoE.

**Evaluation Metrics.** We use Accuracy and FLOPs as evaluation criteria for the classification task.

**Implementation Details.** We construct the experimental models based on the Vision Transformer (ViT) [40]. We follow most of the setups and training configurations of ViT, with specific modifications such as integrating our UoE mechanism into transformer blocks, or replacing them with baseline MoE models. We apply the data selection paradigm to attention blocks and the expert selection paradigm to MLP blocks with similar expert number and activation ratio settings. We designate this experimental group as UoE or UoE-DE. To further explore the influence of the selection mechanism on model performance, we also provide a control group in which data selection is applied to both the attention block and the MLP block, which we denote as UoE-DD.

**Results.** The experimental results are illustrated in Table 3. Since the two tasks share the same model, they exhibit the same FLOPs values. The results show that our model significantly surpasses all state-of-the-art baselines. By introducing Rotary Position Embedding (RoPE) to transformer model design, our UoE achieves a significant accuracy improvement of 8.25% in CIFAR-10 and 12.87% in CIFAR-100 compared to the standard transformer while maintaining the same parameter count and only 64.77% of the FLOPs. In comparison with MoE methods, our model surpasses the well-performing DeepSeek-V3/Skywork-MoE model by 1.19%/1.83% in CIFAR-10 and 1.93%/6.22% in CIFAR-100, which can be attributed to our in-depth expert collaboration and fine-grained dynamic selection on both the attention block and the MLP block.

On the other hand, Skywork-MoE, DeepSeek-MoE, and XMoE primarily concentrate on improving the routing algorithm of MLP experts. However, Skywork-MoE applies the RoPE to effectively capture relative positional information in sequences, while the remaining two models do not conduct significant refinement on the attention block. With RoPE and selective attention optimization, our UoE model outperforms DeepSeek-MoE/XMoE by 10.39%/11.80% on CIFAR-10 and 14.65%/14.20% on CIFAR-100. In terms of computational efficiency, our model achieves 220.34 GFLOPs, matching the efficiency of the most efficient DeepSeek model (210.38 GFLOPs), while significantly surpassing other competitors (best: 245.08 GFLOPs from DeepSeek-MoE). This clearly demonstrates the efficiency-enhancing effect of the selective mechanism on our model.

We also provide the results of the control group UoE-DD. The control group attained accuracies of 81.81 and 55.09 for CIFAR-10 and CIFAR-100, respectively, which is slightly lower than that of the experimental group but superior to all MoE comparison models. We posit that this may be attributed to the complementary effects of the two mechanisms at local and global levels: In the data selection paradigm, attention experts take a subset of the input sequences and provide a lower-level, localized analysis. While in the expert selection paradigm, MLP experts integrate all local results from attention experts and provide the final outcome from a global perspective. In this process, the information within inputs was thoroughly integrated and comprehensively reviewed, leading to enhanced performance.

### *D. Ablation experiments*

Our previous experiments indicate that UoE surpasses state-of-the-art competing models in a variety of tasks. To further verify the positive contribution of UoE's attention and MLP components to our model's state-of-the-art performance, we conducted comparative experiments on both of them with the corresponding components in advanced competing models. The experiments are performed on the Listops and Wikitext-103 datasets, following the experimental design in sections 4A and 4B. For model setup, we adopt the SMHA and UoME in the data selection paradigm with the configuration of $n = 4, r = 0.5$, and utilize the MHA block and the two-layer MLP block in the standard transformer as the basic configuration. For the attention block, we select the MLA block, as implemented in DeepSeek-V2/V3, to serve as our comparison model. For the MLP block, we choose the MoE components of Skywork-MoE, DeepSeek-MoE and XMoE as the comparison model. The computational complexity of the comparison model is on par with that of the corresponding UoE components. The remaining components of the experimental models are configured as the matching components of the standard transformer. The experimental results are shown in Table 4.

TABLE IV
COMPARATIVE EXPERIMENTS OF UOE'S SMHA BLOCK AND UOME BLOCK WITH CORRESPONDING COMPONENTS IN COMPARISON MODELS

| Model | Listops | | Wikitext-103 | |
|---|---|---|---|---|
| | Acc. | GFLOPs | PPL | TFLOPs |
| MLA [31] | 38.16 | 5.48 | 24.29 | 3.01 |
| XMoE [32] | 38.00 | 7.31 | 30.60 | 2.67 |
| DeepSeek-MoE [27] | 37.80 | 7.21 | 28.30 | 2.31 |
| Skywork-MoE [26] | 37.85 | 7.25 | 29.10 | 2.42 |
| SMHA (ours) | 38.41 | **2.99** | **22.68** | 2.55 |
| UoME (ours) | **38.56** | 7.10 | 24.26 | **1.95** |

Our results demonstrate that UoE's attention and MLP blocks significantly outperform the corresponding components of existing state-of-the-art models. With minimal computation expenses, SMHA achieved an accuracy of 38.41 in the Listops task, which is higher than MLA's 38.16. In the Wikitext-103 task, it takes the best-performing perplexity of 22.68. We posit that a significant reason for MLA's effectiveness lies in the indirect increase of model depth facilitated by the introduction of the LoRA mechanism. In UoE, by incorporating the selection and routing mechanism into attention heads, we are able to take advantage of task-related experts while mitigating the impact of unhelpful experts, thereby achieving competitive performance in an alternative manner.

The remaining results further demonstrate the superiority of the UoE architecture. Compared to Skywork-MoE, DeepSeek-MoE, and XMoE, our UoME model only exhibits a significant difference in MLP components. Given the reduced computational cost, the accuracy/perplexity of UoME in the Listops/Wikitext-103 tasks, at 38.56/24.26, is markedly superior to that of Skywork-MoE (37.85/29.10), DeepSeek-MoE (37.80/28.30), and XMoE (38.00/32.60). Within the UoME, the activated experts function as a cohesive unit. The enhanced interaction in experts contributes to additional performance benefits, which emerges as a critical factor in surpassing existing MoE methods.

Then we vary the sequence length and measure the runtime and memory usage of UoE against various comparison models on one A100 GPU with 80 GB HBM. To facilitate a consistent comparison, we maintain the configuration of the context modeling experiment in section 4B, with the only modification being the substitution of the input tensor with randomly generated tensors of varying sequence lengths, similar in format to the original data. We utilize the Pytorch profiler to measure the total time cost of both forward and backward propagation, and the nvtop visualization tool to monitor the peak GPU memory usage during model execution. The maximum sequence length was configured to 4096, as further increases to 8192 would cause a memory overflow.

**Runtime.** Fig. 4 (left) illustrates the runtime in milliseconds of the forward + backward pass of UoE compared to DeepSeek-V3, Skywork-MoE, DeepSeek-MoE, and XMoE. The runtime exhibits a quadratic growth with respect to the sequence length. Benefiting from a more efficient implementation, our UoE model runs significantly faster than all comparison models, except in cases where the input sequence length is less than 1024, the speed is slightly slower than XMoE. We posit that XMoE naturally inherits the optimization strategies from Megatron-LM, thereby reducing the additional overhead caused by non-matrix operations. These optimizations are also applicable to other models, including ours. As the sequence length increases, matrix operations gradually become the dominant factor in the total time cost. In this scenario, our model outperforms all others by virtue of its lower computational complexity. Especially, when increasing the length to 4096, we achieve an average 2.26× speedup relative to the comparison models.

We also note that the models' actual runtime does not strictly correlate with the theoretical FLOPs. Specifically, DeepSeek-V3 and Skywork-MoE perform rather mediocrely in this experiment. Note that our focus is limited to the algorithm itself and does not extend to hardware optimizations. In terms of algorithm implementation, the MoE mechanism in the DeepSeek Series, Skywork-MoE and XMoE is serial, leading to lower computational efficiency than our parallel-optimized UoE model. On the other hand, the specific structures of these comparison models may potentially impact their efficiency. The Multi-Head Latent Attention in DeepSeek-V3 employs the compression mechanism from LoRA, which potentially increases the number of matrix operations. Moreover, when the LoRA rank is relatively small, efficient parallel computation may be compromised. The MoE component of Skywork-MoE conducts serial indexing operations on expert weights, resulting in additional time overhead. By comparison, our model does not exhibit these issues, thereby achieving superior runtime performance.

**Memory Footprint.** Fig. 4 (right) shows the memory footprint of UoE compared to DeepSeek-V3, Skywork-MoE, DeepSeek-MoE and XMoE. In general, the memory footprint grows quadratically with the sequence length. In comparison with competing models, our UoE achieves an average 2.68× memory efficiency improvement. We declare that UoE is inherently superior in terms of memory efficiency. When the attention block and MLP block both activate 50% of the parameters, the theoretical memory efficiency can reach up to 4× greater than that of the fully activated condition, showing a significant advantage over comparison models.

In the experiment, the comparison model employed MoE components with a configuration similar to UoE's MLP block. The MLP experts in DeepSeek Series models consist of three linear layers, which entails higher computational complexity. Moreover, XMoE has a relatively complex overall structure. As a result, DeepSeek-MoE and XMoE exhibit the highest memory usage in the figure. The memory usage of Skywork-MoE's attention block is comparable to that of standard Multi-

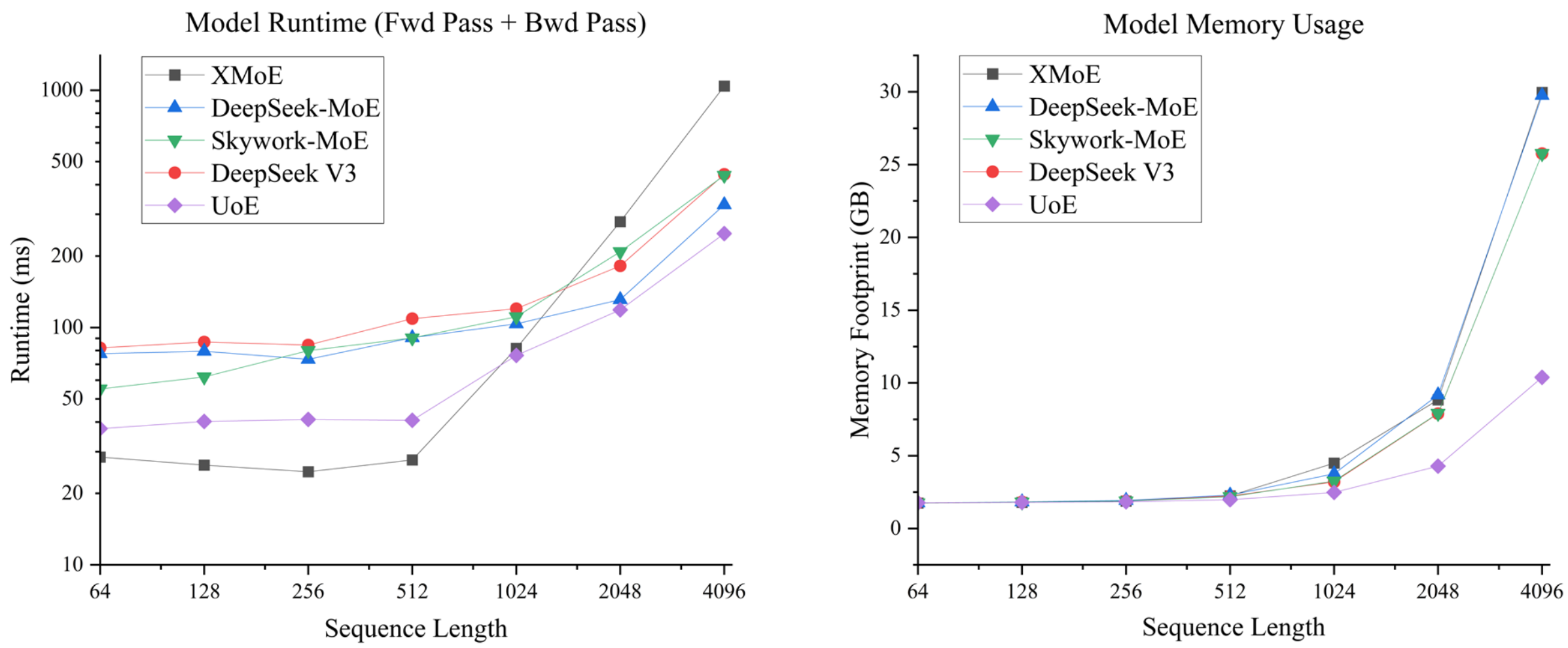

**Fig. 4.** Runtime (left) and memory usage (right) of the experimental model.

Head Attention. DeepSeek-V3 adopts an efficient Multi-Head Latent Attention mechanism, which helps to mitigate the memory efficiency loss from the complex MLP expert. As a result, it demonstrates a memory footprint analogous to Skywork-MoE. It is worth mentioning that in this experiment, the attention block has a greater impact on memory efficiency than the MLP block. With the Selective Multi-Head Attention mechanism, our UoE achieves a more remarkable efficiency improvement. We further conducted ablation experiments on model configurations, with full results and analysis provided in Appendix C.

In summary, we conducted comprehensive and meticulous experiments across multiple representative datasets from both the image and natural language domains. The experimental results have demonstrated that our UoE model surpasses Full Attention models and a group of state-of-the-art MoEs and efficient transformers in several tasks while attaining computational complexity that matches or surpasses that of state-of-the-art models. In addition, the recent experimental results in the released works NSA [19] and MoBA [20] partially and indirectly validate the effectiveness of UoE's SMHA mechanism. Overall, this experiment has provided valuable insights into the effectiveness and efficiency of our UoE model. Further research could focus on the application of UoE in the training and inference of large language models.

## V. Conclusion

We propose the Union-of-Experts, a novel and effective MoE-type framework. To build the UoE architecture, we leveraged a lossless decomposition method to transform an entire transformer model into a group of experts, and developed a hierarchical selection mechanism to dynamically selects task-relevant input parts and allocates them to the activated attention or MLP experts. In this manner, we improve the collective intelligence and knowledge sharing ability of experts, and further enhance the computational efficiency of MoE methods. The experiments on 9 tasks from 3 challenging benchmarks demonstrate that our UoE significantly surpasses state-of-the-art MoE and transformer variants with up to 50% less computational cost, indicating its superior effectiveness and efficiency in various tasks across image and language domains. Considering the extensive applications of MoE and transformers within large language models, our work has promising development and application prospects. For research purposes, we release all the source code of UoE to the public. We aspire for this work to provide valuable insights into the performance and efficiency enhancement of large-scale language models and hope that it will facilitate further research and development.

## References

[1] N. Shazeer *et al.*, "Outrageously large neural networks: The sparsely-gated mixture-of-experts layer," *arXiv preprint arXiv:1701.06538,* 2017.

[2] Z.-H. Zhou, *Ensemble methods: foundations and algorithms*. CRC press, 2012.

[3] W. Li, Y. Peng, M. Zhang, L. Ding, H. Hu, and L. Shen, "Deep model fusion: A survey," *arXiv preprint arXiv:2309.15698,* 2023.

[4] M. Shoeybi, M. Patwary, R. Puri, P. LeGresley, J. Casper, and B. Catanzaro, "Megatron-lm: Training multi-billion parameter language models using model parallelism," *arXiv preprint arXiv:1909.08053,* 2019.

[5] A. Vaswani, "Attention is all you need," *Advances in Neural Information Processing Systems,* 2017.

[6] R. Child, S. Gray, A. Radford, and I. Sutskever, "Generating long sequences with sparse transformers," *arXiv preprint arXiv:1904.10509,* 2019.

[7] S. Wang, B. Z. Li, M. Khabsa, H. Fang, and H. Ma, "Linformer: Self-attention with linear complexity," *arXiv preprint arXiv:2006.04768,* 2020.

[8] Z. Qin *et al.*, "Scaling transnormer to 175 billion parameters," *arXiv preprint arXiv:2307.14995,* 2023.

[9] Z. Qin, W. Sun, D. Li, X. Shen, W. Sun, and Y. Zhong, "Lightning attention-2: A free lunch for handling unlimited sequence lengths in large language models," *arXiv preprint arXiv:2401.04658,* 2024.

[10] N. Kitaev, Ł. Kaiser, and A. Levskaya, "Reformer: The efficient transformer," *arXiv preprint arXiv:2001.04451,* 2020.

[11] T. Dao, D. Fu, S. Ermon, A. Rudra, and C. Ré, "Flashattention: Fast and memory-efficient exact attention with io-awareness," *Advances in neural information processing systems,* vol. 35, pp. 16344-16359, 2022.

[12] T. Dao, "Flashattention-2: Faster attention with better parallelism and work partitioning," *arXiv preprint arXiv:2307.08691,* 2023.

[13] A. Liu *et al.*, "Deepseek-v2: A strong, economical, and efficient mixture-of-experts language model," *arXiv preprint arXiv:2405.04434,* 2024.
[14] Y. Zhang *et al.*, "Tensor Product Attention Is All You Need," *arXiv preprint arXiv:2501.06425,* 2025.
[15] Z. Dai, "Transformer-xl: Attentive language models beyond a fixed-length context," *arXiv preprint arXiv:1901.02860,* 2019.
[16] Y. Tay, D. Bahri, L. Yang, D. Metzler, and D.-C. Juan, "Sparse sinkhorn attention," in *International Conference on Machine Learning*, 2020: PMLR, pp. 9438-9447.
[17] W. Luo, Y. Liu, B. Li, W. Hu, Y. Miao, and Y. Li, "Long-Short Term Cross-Transformer in Compressed Domain for Few-Shot Video Classification," in *IJCAI*, 2022, pp. 1247-1253.
[18] Y. Gao *et al.*, "Seerattention: Learning intrinsic sparse attention in your llms," *arXiv preprint arXiv:2410.13276,* 2024.
[19] J. Yuan *et al.*, "Native Sparse Attention: Hardware-Aligned and Natively Trainable Sparse Attention," *arXiv preprint arXiv:2502.11089,* 2025.
[20] E. Lu *et al.*, "MoBA: Mixture of Block Attention for Long-Context LLMs," *arXiv preprint arXiv:2502.13189,* 2025.
[21] W. Fedus, J. Dean, and B. Zoph, "A review of sparse expert models in deep learning," *arXiv preprint arXiv:2209.01667,* 2022.
[22] D. Lepikhin *et al.*, "Gshard: Scaling giant models with conditional computation and automatic sharding," *arXiv preprint arXiv:2006.16668,* 2020.
[23] W. Fedus, B. Zoph, and N. Shazeer, "Switch transformers: Scaling to trillion parameter models with simple and efficient sparsity," *Journal of Machine Learning Research,* vol. 23, no. 120, pp. 1-39, 2022.
[24] S. Gururangan, M. Lewis, A. Holtzman, N. A. Smith, and L. Zettlemoyer, "Demix layers: Disentangling domains for modular language modeling," *arXiv preprint arXiv:2108.05036,* 2021.
[25] J. Wang, J. Wang, B. Athiwaratkun, C. Zhang, and J. Zou, "Mixture-of-Agents Enhances Large Language Model Capabilities," *arXiv preprint arXiv:2406.04692,* 2024.
[26] T. Wei *et al.*, "Skywork-moe: A deep dive into training techniques for mixture-of-experts language models," *arXiv preprint arXiv:2406.06563,* 2024.
[27] D. Dai *et al.*, "Deepseekmoe: Towards ultimate expert specialization in mixture-of-experts language models," *arXiv preprint arXiv:2401.06066,* 2024.
[28] S. Moreno-Alvarez, J. M. Haut, M. E. Paoletti, and J. A. Rico-Gallego, "Heterogeneous model parallelism for deep neural networks," *Neurocomputing,* vol. 441, pp. 1-12, 2021.
[29] Y. Huang *et al.*, "Gpipe: Efficient training of giant neural networks using pipeline parallelism," *Advances in neural information processing systems,* vol. 32, 2019.
[30] L. Wang, H. Gao, C. Zhao, X. Sun, and D. Dai, "Auxiliary-loss-free load balancing strategy for mixture-of-experts," *arXiv preprint arXiv:2408.15664,* 2024.
[31] A. Liu *et al.*, "Deepseek-v3 technical report," *arXiv preprint arXiv:2412.19437,* 2024.
[32] Y. Yang, S. Qi, W. Gu, C. Wang, C. Gao, and Z. Xu, "XMoE: Sparse Models with Fine-grained and Adaptive Expert Selection," in *Findings of the Association for Computational Linguistics ACL 2024*, 2024, pp. 11664-11674.
[33] R. Al-Rfou, D. Choe, N. Constant, M. Guo, and L. Jones, "Character-level language modeling with deeper self-attention," in *Proceedings of the AAAI conference on artificial intelligence*, 2019, vol. 33, no. 01, pp. 3159-3166.
[34] Y. Ju. "calflops: a FLOPs and Params calculate tool for neural networks." https://github.com/MrYxJ/calculate-flops.pytorch (accessed Jul. 24, 2024).
[35] Y. Tay *et al.*, "Long range arena: A benchmark for efficient transformers," *arXiv preprint arXiv:2011.04006,* 2020.
[36] K. Choromanski *et al.*, "Rethinking attention with performers," *arXiv preprint arXiv:2009.14794,* 2020.
[37] A. Katharopoulos, A. Vyas, N. Pappas, and F. Fleuret, "Transformers are rnns: Fast autoregressive transformers with linear attention," in *International conference on machine learning*, 2020: PMLR, pp. 5156-5165.
[38] Y. Xiong *et al.*, "Nyströmformer: A nyström-based algorithm for approximating self-attention," in *Proceedings of the AAAI Conference on Artificial Intelligence*, 2021, vol. 35, no. 16, pp. 14138-14148.
[39] C. Zhu *et al.*, "Long-short transformer: Efficient transformers for language and vision," *Advances in neural information processing systems,* vol. 34, pp. 17723-17736, 2021.
[40] A. Dosovitskiy, "An image is worth 16x16 words: Transformers for image recognition at scale," *arXiv preprint arXiv:2010.11929,* 2020.

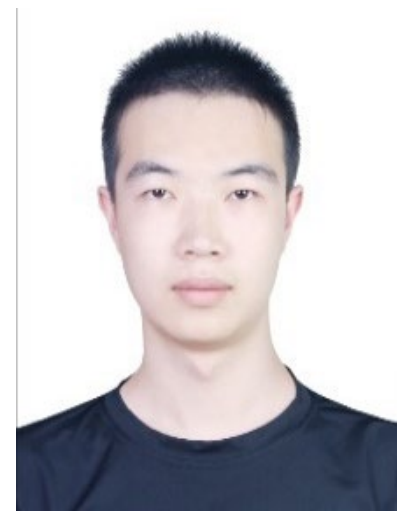

**Yujiao Yang** received the B.E. degree in vehicle engineering from Nanjing Tech University, Nanjing, China. He is currently pursuing the M.E. degree with the Dalian University of Technology, Dalian, China. His research interests include foundation models, large-scale models for autonomous driving, and generative models.

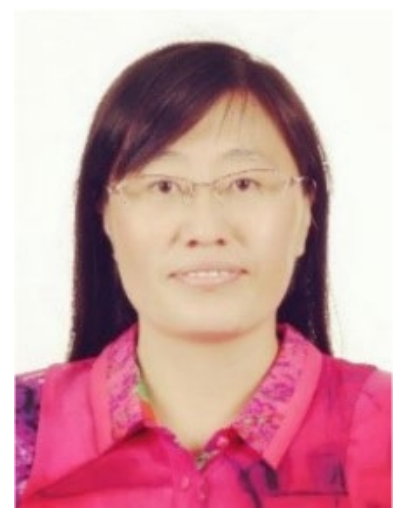

**Jing Lian** (Member, IEEE) received the Ph.D. degree in communication and information system from Jilin University, Jilin, China, in 2008. She is currently an Associate Professor and the Deputy Director of the Automotive Electronic Institute, Dalian University of Technology and a Judicial expert in the vehicle performance. She is the leader of more than 20 research projects. She is the author of over 70 publications. Her research interests include environmental perception, automotive electronics, trajectory prediction, and control of intelligent vehicle.

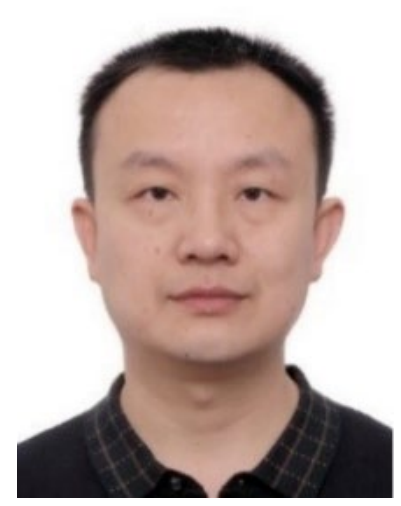

**Linhui Li** (Member, IEEE) received the Ph.D. degree in vehicle operation engineering from Jilin University, Jilin, China, in 2008. He is an associate professor at Dalian University of Technology. He was a visiting scholar at The Ohio State University from 2017 to 2018, and is the author of over 40 publications. His main research interests include intelligent vehicle trajectory prediction, vision based environmental perception of intelligent vehicle, and navigation control.

## APPENDICES

### A. Lossless Decomposition of Transformer

Here we detail the implementation of splitting a transformer model into an equivalent set of experts. A classical transformer can be comprised of two main components: the Multi-Head Attention (MHA) block and the Multi-Layer Perceptron (MLP) block.

**1. MLP Block**

We start by detailing the MLP block. We referenced the method introduced by Shoeybi et al. [4] to decompose a standard two-layer MLP into $n$ experts. WLOG, let $n=2$. A two-layer MLP is a combination of two linear layers where each layer's output $Y$ is calculated through a multiplication of the input batch $X$ and the transformation matrix $A$ followed by an activation function $\varphi(\cdot)$: $Y=\varphi(XA)$. We can make an equivalent decomposition on a linear layer through two methods. One is to split A along its column dimension, i.e., $A=[A_1,A_2]$. In this pattern, the activation function can be applied to the output of each partitioned matrix multiplication independently. The result $Y$ can be calculated as follows:

$$Y=\varphi(X\left[A_1,A_2\right])=[\varphi(XA_1),\varphi(XA_2)] \tag{24}$$

The equation can be further categorized into a broadcast operation and an element-wise multiplication with activation:

$$X'=\left[X,X\right]=\left[X_1',X_2'\right] \tag{25}$$

$$\begin{aligned}Y&=\varphi\left(\left[X_1',X_2'\right]\odot\left[A_1,A_2\right]\right)\\&=\left[\varphi\left(X_1'A_1\right),\varphi\left(X_2'A_2\right)\right]\end{aligned} \tag{26}$$

Where $\odot$ denotes the Hadamard product. Another method is to split $A$ along its row dimension and $X$ along its column dimension, as shown in Equation 4:

$$X=[X_1,X_2],A=\begin{bmatrix}A_1\\A_2\end{bmatrix} \tag{27}$$

With this pattern, we compute the result $Y$ as follows:

$$Y=\varphi\left(\left[X_1,X_2\right]\begin{bmatrix}A_1\\A_2\end{bmatrix}\right)=\varphi\left(X_1A_1+X_2A_2\right) \tag{28}$$

Similarly, it can be decomposed into an element-wise multiplication and a reduce operation with activation:

$$\begin{aligned}Y'&=\left[X_1,X_2\right]\odot\left[A_1,A_2\right]\\&=\left[X_1A_1,X_2A_2\right]=\left[Y_1',Y_2'\right]\end{aligned} \tag{29}$$

$$Y=\varphi\left(\sum_{i=1}^{2}Y_i'\right)=\varphi\left(Y_1'+Y_2'\right) \tag{30}$$

Note that due to the nonlinearity of the activation function $\varphi(\cdot)$, $\varphi\left(Y_1'+Y_2'\right)\neq\varphi\left(Y_1'\right)+\varphi\left(Y_2'\right)$. When incorporating the two decomposition procedures into our method, we employ an index select function in place of the broadcast operation in Equation 25, and replace the sum function in Equation 30 with the index add function. Please refer to sections B and C for more details.

Within one expert, the computation process should operate independently, which means that synchronization operations such as broadcast and reduce cannot be performed during computation. Inspired by Shoeybi et al. [4], we split the MLP's first layer in the column dimension and partition the second one in the row dimension. In this way, the synchronizations only occur at the beginning and ending of the MLP's calculation. Thus, we can take each independent part as an expert.

**2. MHA Block**

Then we turn to a more complex scenario: the attention block. An attention block can be decomposed into four steps:

$$Q,K,V=XW^I \tag{31}$$

$$S=\theta\left(Q^TK\right) \tag{32}$$

$$O=SV \tag{33}$$

$$Y=OW^O \tag{34}$$

Where $X$ and $Y$ denote the input and output batch. $Q$, $K$ and $V$ denote the query, key and value. $W^I$ and $W^O$ denote the input and output projection matrices. $\theta(\cdot)$ represents a nonlinear transformation function. Similar to the MLP situation, we apply $n$ expert column partition on $W^I$, and the corresponding row partition on $W^O$. Thus, we acquire a group of $n$ independent experts, which as a whole is equivalent to $n$ head Attention block.

### B. Dynamic Routing Strategy of Experts

In this section, we introduce our unique hierarchical routing method. Fig. 5 presents this processing workflow. Generally, our proposed routing method comprises two modes, which perform a division-selection mechanism separately on samples and models. The two modes can be implemented separately in different blocks, as well as simultaneously enabled within one block, although the latter may incur additional computation costs.

**1. Data Selection Mode**

At each propagation step, the UoE takes an input batch $X$. We can improve efficiency by removing input fragments that are predicted to have negligible impact on each expert's computation. Let $d$ denote the length of the token dimension. For a single sample $x\in\mathbb{R}^{l\times d}$ within a batch $X$ of $b$ samples, we split them into $m$ patches with length $l_p$ along the sequence dimension:

$$\begin{aligned}x'&=\left[p_0,\dots p_m\right]\\&=\left[\underbrace{\overbrace{x_0,\dots x_{l_p-1}}^{l_p},\dots\overbrace{x_{(m-1)l_p},\dots x_{ml_p-1}}^{l_p}}_{l_p\times m}\right]\end{aligned} \tag{35}$$

Where $x'\in\mathbb{R}^{m\times l_p\times d}$. We route each patch to the best determined top-k experts, selected from a set of $n$ experts. The gating score $g'\in\mathbb{R}^{n\times m}$ is streamlined by a two-layer feed-forward network (FFN):

$$g'=\mathrm{FFN}_A(\mathrm{FFN}_B(x)^T) \tag{36}$$

We apply a softmax activation to calculate the gating value $g_{i,j}$ of the j-th patch $p_j$ with respect to the i-th expert $E_i$:

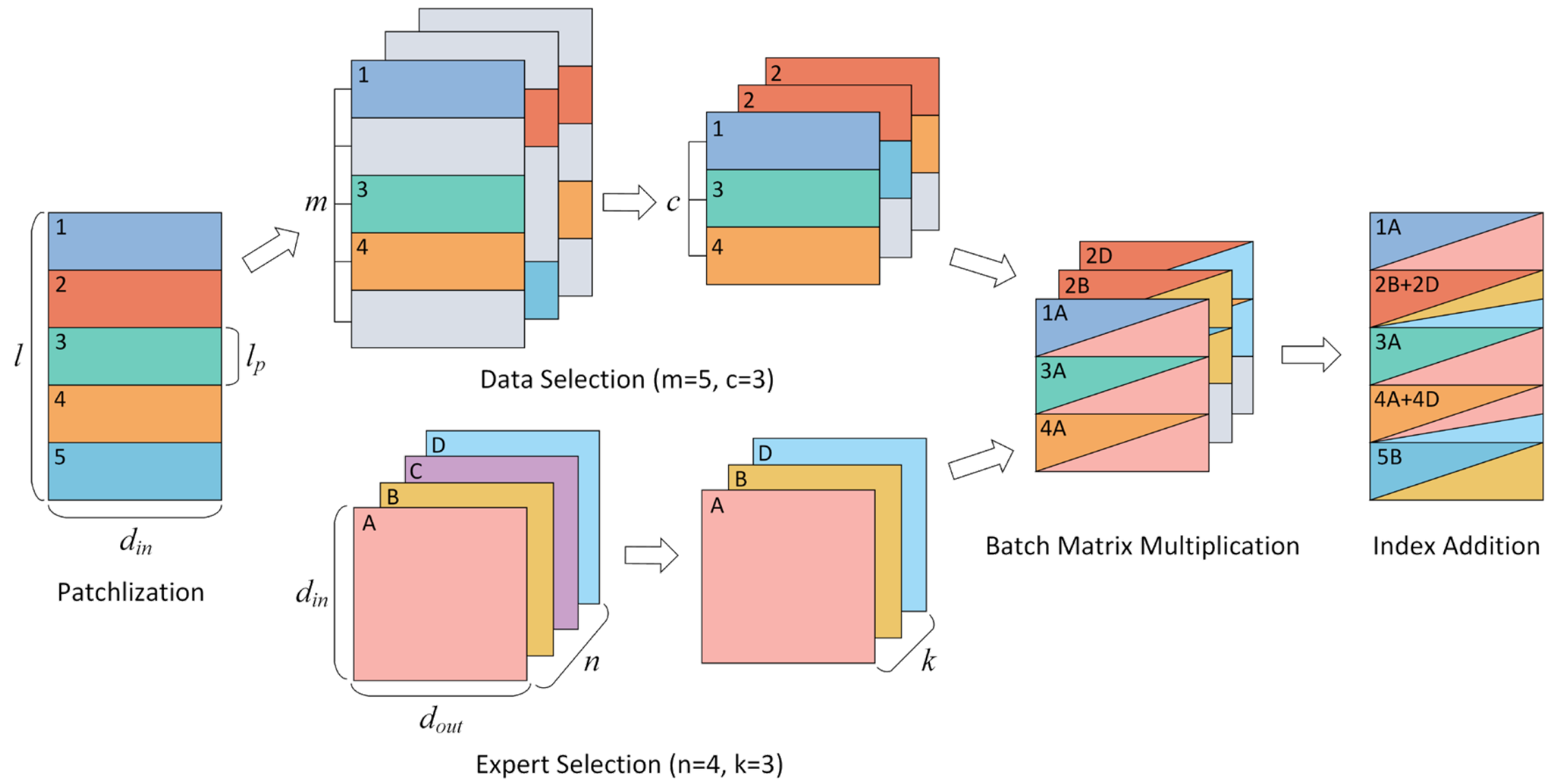

**Fig. 5.** The processing workflow of UoE's routing mechanism. Initially, we split each input sample into $m$ patches and decompose a whole model into $n$ experts. In the data selection paradigm, we select no more than $c$ patches for each expert as input. In the expert selection paradigm, we route each original or selected input to the top $k$ experts. We compute the output of experts in parallel and aggregate the results to the correct positions of the final output.

$$g_{i,j} = \frac{\exp(g'_{i,j})}{\sum_{t=0}^{n} \exp(g'_{t,j})} \tag{37}$$

For each expert, we select a portion of patches as the input. The strategy that simply selects the top-k patches for each expert is unable to dynamically adjust the amounts of valid patches based on input type and task complexity, leading to an inappropriate discarding of beneficial patches. Regarding this matter, we propose a two-stage routing method to achieve dynamic allocation of input patches. First, we route each patch to $k$ experts with the highest probabilities. The indexes of experts $id' \in \mathbb{R}^{m \times k}$ with respect to the j-th patches can be obtained by implementing a topk function:

$$id'_j = \mathrm{Topk}(\{g_{i,j} \,|1 \le i \le n\}, k) \tag{38}$$

We then calculate the maximum counts $c$ of patches received by every expert from each sample:

$$c = \max_{i=1}^{n}(\sum_{j=1}^{m} I(i \in id'_j)) \tag{39}$$

Where $I$ denotes the indicator function. $I_{(i \in id_j')} = 1$ if $i$ is in $id'_j$, and 0 otherwise. We select the $c$ patches with the highest gating value as each expert's input. The index of the patches $id \in \mathbb{R}^{n \times c}$ that are routed to the i-th expert is calculated as follows:

$$id_i = \mathrm{Topk}(\{g_{i,j} \,|1 \le j \le m\}, c) \tag{40}$$

We sort the indices to restore their original relative position. Based on the indices, we extract the patches each expert requires. This step is implemented through an index selection function $f_s$:

$$x''_i = f_S(x', id_i) \tag{41}$$

Where $x'' \in \mathbb{R}^{n \times c \times l_p \times d}$. In this manner, each expert receives a copy of a group of patches as input, as shown in Fig. 2(c). The above processing maintains the relative positions of patches while altering their absolute positions. However, in our method, the Rotary Position Embedding (RoPE) positional is added at the beginning of each attention block, so this change of absolute positions will not result in the loss of location information. See the next section for more details.

Then we process each input $x''$ with the i-th expert $E_i$, which is implemented in parallel through batch matrix operations:

$$y'_i = E_i(x''_i) \tag{42}$$

Note that each expert $E_i$ here refers to one part of a whole union, $y' \in \mathbb{R}^{n \times c \times l_p \times d}$. This kind of union can not only be like a dense model, but also be like a multi-head one. Specifically, if $E_i$ is linear, we get a union like a dense model; otherwise, the union will be like a multi-head one. In order to derive the final output, we apply an index add function to gather $y'$ into the original input sample $x$. For the j-th output patch $y_j$, the procedure is described as follows:

$$y_j = x_j + \sum_{i=1}^{n} \sum_{t=0}^{c} y'_{i,t} \cdot I(id_{i,t} = j) \tag{43}$$

Where $y \in \mathbb{R}^{m \times l_p \times d} \xrightarrow{reshape} \mathbb{R}^{l \times d}$ and $I_{j \in id_i}$ denotes the indicator function. We add outputs of selected patches to their corresponding location in the input tensor. In other words, we apply a selective residual connection to form the output tensor. This approach not only realizes fine-grained routing but also preserves the locality within patches, thereby effectively improving UoE's performance.

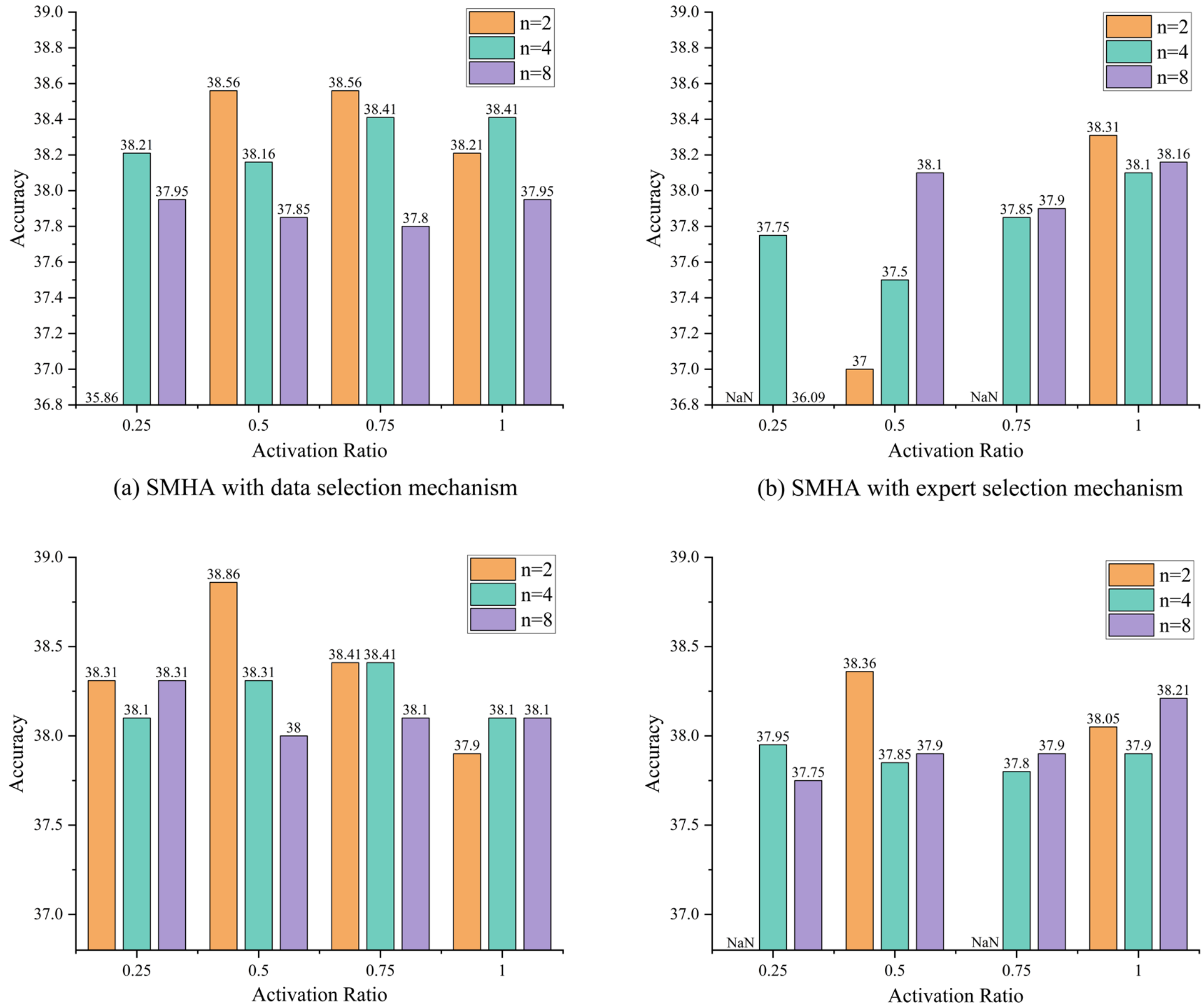

(a) SMHA with data selection mechanism (b) SMHA with expert selection mechanism

(c) UoME with data selection mechanism (d) UoME with expert selection mechanism

**Fig. 6.** Ablation study on the number of experts $n$ and the activation ratio $r$.

**2. Expert Selection Mode**

In expert selection mode, we improve efficiency by limiting the number of activated experts. Note that compared to data selection, this method employs the batch rather than the sample as the processing subject. To initiate this process, we take a batch of samples $X$ as input, $X \in \mathbb{R}^{b \times l \times d} \xrightarrow{reshape} X' \in \mathbb{R}^{b \times (l \times d)}$. The gating value $g \in \mathbb{R}^{b \times k}$ can be calculated as follows:

$$g = \mathrm{softmax}(\mathrm{FFN}(X')) \tag{44}$$

We route the j-th sample in the batch $X$ to the optimal $k$ of $i$ experts, and the routing indices $id' \in \mathbb{R}^{b \times k}$ of the j-th sample can be obtained through a topk function:

$$id'_j = \mathrm{Topk}(\{g_{i,j} \,|\, 1 \le i \le n\}, k) \tag{45}$$

The sample indices tuple $id$ belonging to the i-th expert can be denoted as follows:

$$id^{(i)} = \{id'_j \,|\, j \in \{1, 2, ...m\}, i \in id'_j\} \tag{46}$$

With the sample indices, we apply an index selection function $f_s$ to get the i-th expert's input samples from the original input batch $Y$:

$$X''^{(i)} = f_s(X, id^{(i)}) \tag{47}$$

We enable experts to process their respective received samples:

$$Y'^{(i)} = E_i\left(X''^{(i)}\right) \tag{48}$$

Finally, we gather the outputs into the original input $X$ through an index add function to obtain the final output $Y$. For each output sample $Y_j$, the procedure is described as follows:

$$Y_j = X_j + \sum_{i=1}^{n} \sum_{t=0}^{c_i} Y_t'^{(i)} \cdot I(id_t^{(i)} = j) \tag{49}$$

Where $c_i$ denotes the cardinality of the i-th set of $Y'$ tuple. In comparison with data selection, the input sample in expert selection mode maintains its integrity and continuity across the sequence dimension, which contributes to the robustness enhancement.

We claim that UoE's routing mechanism is essentially a superposition of data selection and expert selection. It is evident that both selection modes are special cases of this paradigm. If both modes are activated within one block, only the selected patches routed to the activated experts will be processed, which enables a fine-grained and flexible routing process. Although it shows considerable potential, its relatively intricate structure imposes extra time costs under the current parallel framework. In practice, we primarily employ a single data selection or expert selection mechanism to construct the model. This contributes to a comprehensive optimization of performance and efficiency.

*C. Ablation Study on model configuration*

Here we implement an ablation study on model configuration, which is depends on two additional parameters introduced in UoE: the number of experts $n$ and the activation ratio of data or experts $r$. The experiments are performed on the Listops dataset, following the experimental design used in section 4B. We conduct experiments with various parameter combinations to evaluate their impact on the two basic components: the Selective Multi-Head Attention (SMHA) block, which symbolizes UoE in a dense condition, and the Union-of-MLP-Experts (UoME) block, which represents that in a multi-head condition. Specifically, we configure two experimental models: the model that adopts the SMHA block and a standard two-layer MLP block, denoted as SMHA, and the model that uses a standard Multi-Head Attention block and a UoME block, denoted as UoME. Fig. 6 illustrates the results of our experiment.

To evaluate the effects of the activation ratio, we set $n = 2,4,8$ and regulated the parameter $r$. As shown in Fig. 6, the variations in activation ratio have no regular pattern of impact on accuracy, which comprehensively showcases the characteristics of MoE-based methods. In MoE frameworks, expert roles are highly differentiated, which means that only a few experts are typically required to effectively process an individual instance taken from a wide-ranging distribution. In our experiments, when $r = 0.25$, the model exhibits suboptimal performance in some conditions. When $r \geq 0.5$, both types of SMHA and UoME models exhibited sufficient fitting capability. Therefore, additional increments in the activation ratio may introduce redundancy and contribute little to enhancing performance. Actually, when increasing the expert count or the number of patches processed by each expert (i.e., increasing $r$), an improvement in training accuracy was observed, indicating slight overfitting in the models. By increasing the dataset size, redundant experts can be transformed into useful experts, which may in turn enhance the model performance.

Then we explore the effects of the number of experts $n$. UoE's attention block integrates the outputs of activated individual attention heads into a comprehensive final outcome. Empirical studies have demonstrated that increasing the number of attention heads facilitates the simultaneous learning of input sequences within distinct subspaces. On the other hand, it may limit the size of the heads, potentially undermining the ability of individual heads to discern subtle details. As illustrated in Table 5, when $n = 2,4$, the data selection version achieves an average accuracy of 38.44/38.30, reflecting the balance between two interacting factors. However, when increasing head counts to 8, the average accuracy decreases to 37.89, indicating that the limited expert scale has led to a considerable degradation in performance. By contrast, the expert selection version achieved a more stable average score of 37.65/37.80/37.56, indicating that it is less affected by the number and size of attention experts, thus demonstrating better robustness.

TABLE V
THE AVERAGE PERFORMANCE OF UOE'S COMPONENTS UNDER DIFFERENT ACTIVATION RATIO FOR N=2, 4 AND 8

| Expert Counts | SMHA | | UoME | |
|---|---|---|---|---|
| | DS. | ES. | DS. | ES. |
| n=2 | 38.44 | 37.65 | 38.37 | 38.21 |
| n=4 | 38.30 | 37.80 | 38.23 | 37.88 |
| n=8 | 37.89 | 37.56 | 38.13 | 37.94 |
| Avg. | 38.21 | 37.67 | 38.24 | 38.01 |

In a UoE style MLP block, activated experts function as an integral part of a dense MLP model, which ensures the model's robustness against the reduction in expert size. As presented in Table 5, when $n$ is varied to 2, 4, and 8, the performance of the data/expert selection UoME model under different activation ratios remains relatively stable, with average values of 38.37/38.21, 38.23/37.88, and 38.13/37.94. When $r < 1$, increasing expert counts leads to a fine-grained and highly specialized expert operation, which is potentially beneficial to performance improvement. However, as illustrated in section 3D, it may induce an increase in computation and memory cost because the expert is too small to load in a CUDA core, thereby wasting unused computation space. In general, we recommend a setting of MLP expert counts that is larger while keeping the expert dimension larger than the size of a CUDA core.